\documentclass[final,12pt]{elsarticle}

\usepackage{graphicx}
\usepackage{multirow}
\usepackage{subcaption}
\usepackage{graphicx}
\usepackage{float} 
\usepackage{booktabs}
\usepackage{tabularx}
\usepackage{xcolor}
\usepackage{caption}
\usepackage{amsmath}
\usepackage{amssymb}
\usepackage{lipsum}

\journal{Journal of Computational Science}

\begin{document}

\begin{frontmatter}





\title{Federated Learning in Genetics: Extended Analysis of Accuracy, Performance and Privacy Trade-offs}








\author[label1,label2]{Anika Hannemann}
\author[label2]{Jan Ewald}
\author[label2]{Leo Seeger}
\author[label1,label2]{Erik Buchmann}
\affiliation[label1]{organization={Dept. of Computer Science}, country={Leipzig University}}

\affiliation[label2]{organization={Center for Scalable Data Analytics and Artificial Intelligence (ScaDS.AI) Dresden/Leipzig, Leipzig University, Germany}, country={ surname@cs.uni-leipzig.de}}

\begin{abstract}
Machine learning on large-scale genomic or transcriptomic data is important for many novel health applications. For example, precision medicine tailors medical treatments to patients on the basis of individual biomarkers, cellular and molecular states, etc. 
However, the data required is sensitive, voluminous, heterogeneous, and typically distributed across locations where dedicated machine learning hardware is not available. Due to privacy and regulatory reasons, it is also problematic to aggregate all data at a trusted third party.
Federated learning is a promising solution to this dilemma, because it enables decentralized, collaborative machine learning without exchanging raw data. 

In this paper, we perform comparative experiments with the federated learning frameworks TensorFlow Federated and Flower. Our test case is the training of disease prognosis and cell type classification models. We train the models with distributed transcriptomic data, considering  both data heterogeneity and architectural heterogeneity. We measure model quality, robustness against privacy-enhancing noise and computational performance. We evaluate the resource overhead of a federated system from both client and global perspectives and assess benefits and limitations. Each of the federated learning frameworks has different strengths. However, our experiments confirm that both frameworks can readily build models on transcriptomic data, without transferring personal raw data to a third party with abundant computational resources. This paper is the extended version of \cite{hannemann2024federated}.
\end{abstract}


\end{frontmatter}

\section{Introduction}
\label{sec:intro}

Machine learning has the potential for a paradigm shift in healthcare, 
towards medical treatments based on individual patient characteristics \cite{wu2022integrating,danek2023federated}. For example, precision medicine uses biomarkers, genome, cellular and molecular data, and considers the environment and the lifestyle of patients \cite{hodson2016precision,kosorok2019precision}. 
This is particularly important in the context of complex, heterogeneous and multi-factorial diseases like cancer, where patients often exhibit great variability of responses to standard treatments. 
Machine learning on large scale genomic and transcriptomic data enables 
the identification of disease subtypes, prediction of disease progression and selection of targeted therapies. Therefore, models need to be trained on large, diverse patient cohorts (sample size) with high-resolution genetic characterization (number of features). This is challenging:  
The data is commonly distributed across multiple healthcare institutions that may not possess the high-performance computing resources needed to build large deep-learning models. The sensitive nature of genomic and transcriptomic data 
presents privacy challenges. Genomic mutations and markers could even allow to re-identify individuals and their relatives~\cite{oestreich2021privacy}. This disallows to freely share such data with centralized aggregators.

\begin{figure}[t]
    \vskip -0.1in
    \centering
    \includegraphics[width=\linewidth, trim={30 20 45 10}, clip]{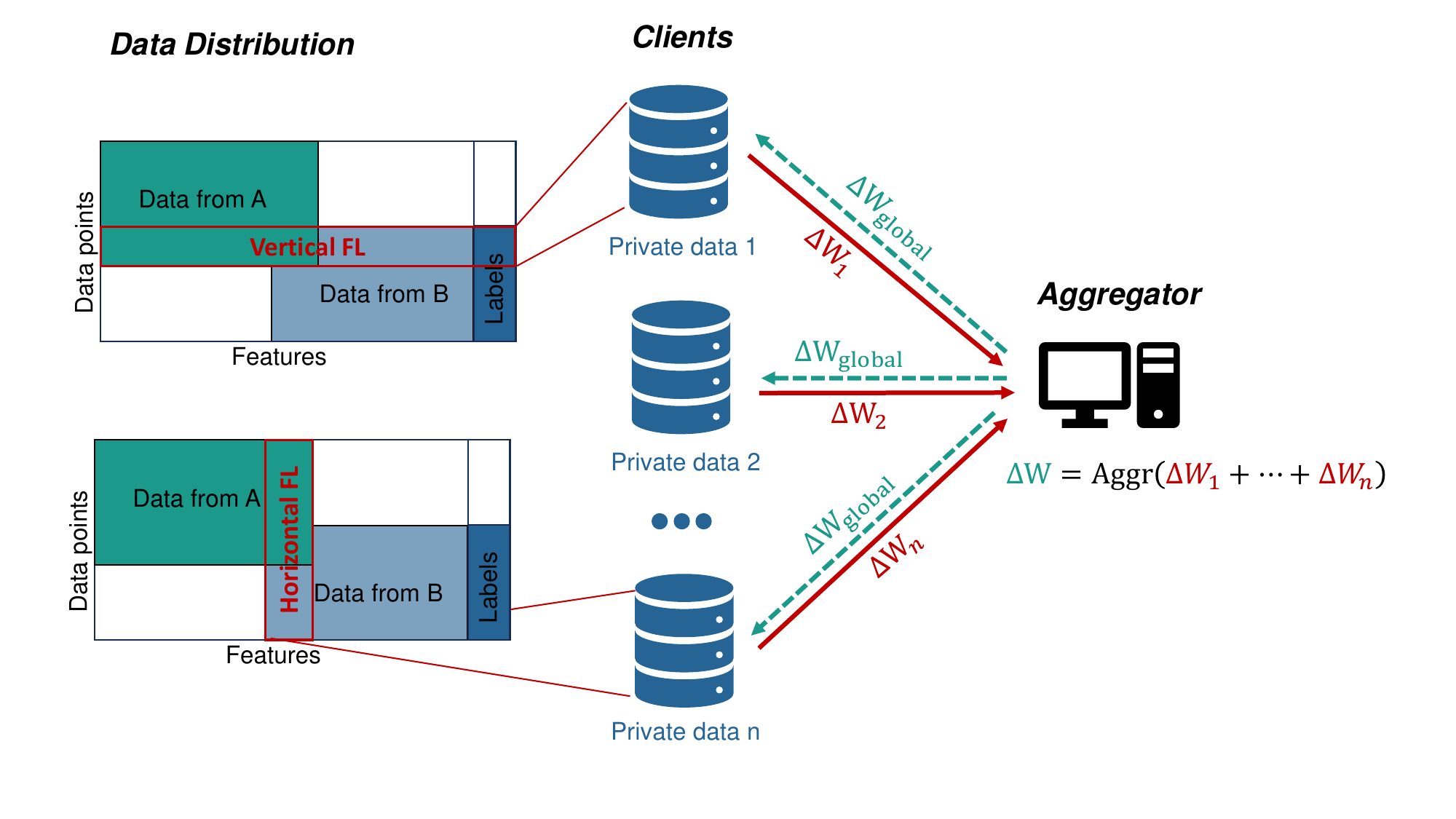}
     \vskip -0.1in
    \caption{Overview of horizontal and vertical federated learning.}
    \label{fig:federated_overview}
\end{figure}

Federated learning (FL), as shown in Figure~\ref{fig:federated_overview}, 
allows for decentralized model training across multiple data sets without requiring to transfer sensitive raw data~\cite{mcmahan2017communication,pfitzner2021federated}. 
Only model parameters are exchanged between the aggregator and participating clients, and the overall computational burden is effectively shared. Adding noise~\cite{adnan2022federated,beguier2021differentially,choudhury2019differential} to shared parameters could further increase privacy.

\begin{figure*}[t]
    \centering
    \includegraphics[width=1\linewidth, trim={70 200 90 0}, clip]{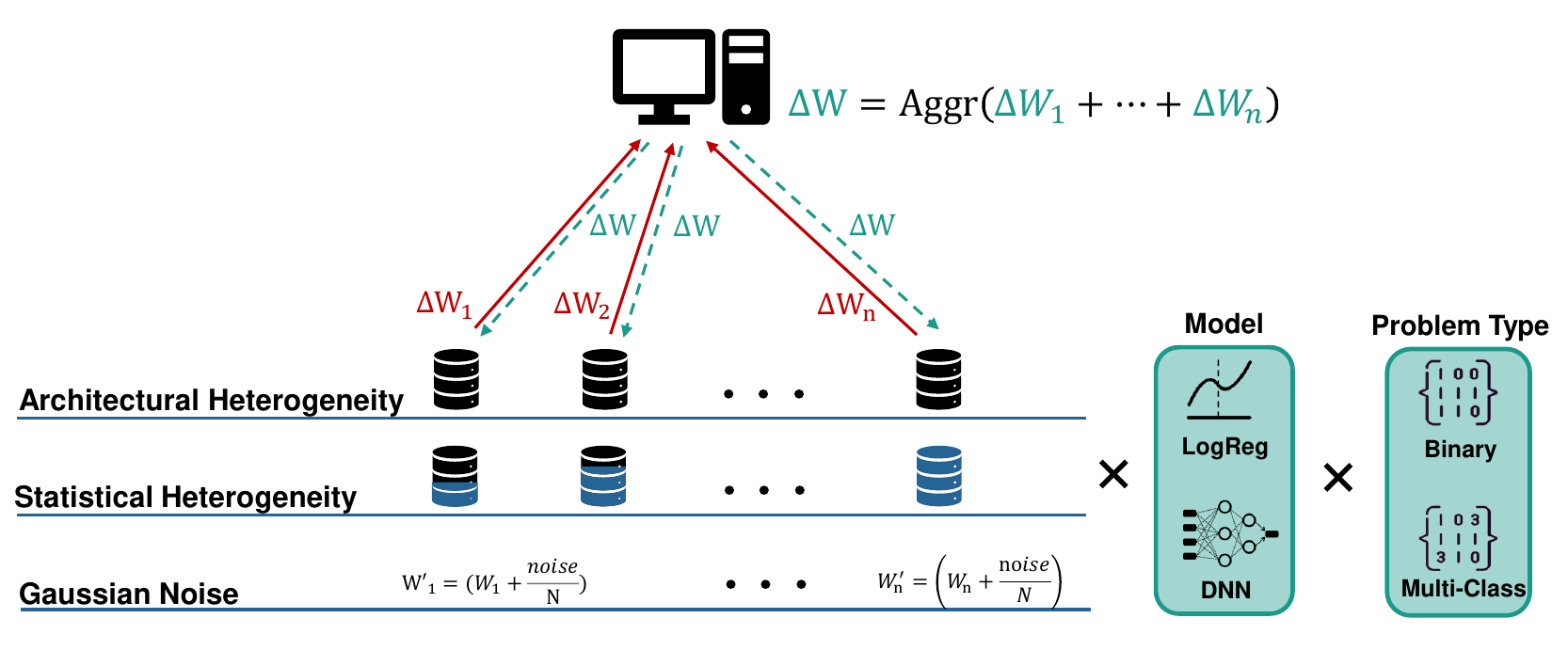}
    \vskip 0.1in
    \caption{Key challenges of federated learning.}  
    \label{fig:overview}
\end{figure*}

In this paper, we investigate the technical and conceptual challenges that arise when implementing FL on transcriptomic data. 
Figure~\ref{fig:overview} illustrates our four key challenges: 
\textit{Architectural Heterogeneity} refers to different numbers of clients with varying computational capabilities. 
\textit{Statistical Heterogeneity} relates to data distributions and sizes. 
\textit{Gaussian Noise} addresses the impact of applying noise to the data, e.g., to achieve Differential Privacy, with different models (Logistic Regression and Sequential Deep Learning) and problem types (Binary and Multi Label).
Finally, \textit{Resource Consumption} addresses storage, communication overhead and training times. 
To explore these challenges, we have conducted comparative experiments with two state-of-the-art FL frameworks -- TensorFlow Federated (TFF) and Flower (FLWR) -- and transcriptomic data. The experiments described in this paper constitute an extended version of those presented in \cite{hannemann2024federated}, incorporating additional results and analyses to build upon the initial findings. In particular, we make the following contributions: 

\begin{itemize}
    \item We train disease prognosis and cell type classification models with TFF and FLWR using hyperparameter tuning, and we measure the model quality. 
    \item We analyze the effects of the number of clients, the amount of training data and data distribution on the global model quality. 
    \item We measure the impact of Gaussian noise, locally applied to the weights, on the global model quality.
    \item We compare memory consumption, run-times and network traffic of TFF and FLWR for different settings.
    \item We examine the benefits and limitations of a federated system from both the client and  global perspective.
\end{itemize}

We have demonstrated that FL frameworks can be readily applied to precision medicine applications. 
Even more, we obtained an excellent global model with an AUC of up to 0.98 for disease prognosis and cell type classification with transcriptomic data. This performance is robust in the presence of diminishing data quality, increasing clients and diverse data distributions, and it reduces the necessary computational resources for the individual medical institution.  


\textbf{Paper structure}: Section \ref{sec:background} introduces background information on federated learning. Section \ref{sec:related} provides a review of related work in the fields of precision medicine, FL applications and frameworks.  
Section~\ref{sec:method} describes our methodology, including data preparation, model architecture, and evaluation metrics. 
Section~\ref{sec:experiments} presents our experimental results. Section \ref{sec:discussion} discusses the results and section~\ref{sec:conclusion} concludes.
 
\newpage

\section{Background}
\label{sec:background}

\subsubsection{Federated learning} \label{FL_background}

Federated learning, introduced by McMahan \textit{et al.} \cite{mcmahan2017communication}, enables users to leverage models on distributed nodes containing rich yet sensitive data. Unlike conventional machine learning, where a model \( f(x; \theta) \) is trained on \( D_{\text{cent}} \), FL addresses privacy concerns or resource limitations by allowing nodes $\mathcal{N}_1, ..., \mathcal{N}_n$ to collaboratively train a model \( f(x; \theta) \)  without sharing their data. In FL, each node $\mathcal{N}_i$ trains a local model $f(x; \theta)_i$ on its data set $\mathcal{D}_i$ and shares the model parameters with a central server. The central server aggregates these parameters to create a global model $f(x; \theta)_{fed}$ with an accuracy of $acc_{fed}$. There are several aggregation algorithms, with FedAvg being the most widely used aggregation method, performing a weighted average of local model updates \cite{mcmahan2017communication}. Given local model parameters \( \theta_i^{(t)} \) from node \( \mathcal{N}_i \) at training round \( t \), the global model update is:  
\begin{equation}
\theta^{(t+1)} = \sum_{i=1}^{n} \frac{|\mathcal{D}_i|}{\sum_{j=1}^{n} |\mathcal{D}_j|} \theta_i^{(t)}
\end{equation}  
where \( |\mathcal{D}_i| \) is the number of data samples at node \( \mathcal{N}_i \). This method ensures that models trained on larger data sets contribute more to the global model.  





As more data is collected, the process is repeated, with each node updating its local model and forwarding the updated parameters to the central server. Thus, the data does not leave its origin at any time in the computation. At some point in the iteration of FL, if $\vert acc_{fed} - acc_{cent} \vert \leq \delta$, then the federated learning framework is said to have $\delta$-accuracy loss. The goal in FL is to have less accuracy loss while maintaining efficiency and the data's privacy. \\

Federated learning, initially developed for edge computing in a \textit{cross-device} context, has garnered interest for its application in broader settings like across multiple institutions, termed \textit{cross-silo}. In this scenario, the focus is on a limited number of reliable participants, such as hospitals, health insurance providers or biomedical research institutions. Another distinction can be made regarding the training data which can be distributed either \textit{horizontally} or \textit{vertically} \cite{kairouz2021advances}. Horizontally distributed data involves data from different organizations or devices with similar feature spaces (e.g. human genes) but different samples (e.g. patients), while vertically distributed data occurs when institutions have different feature spaces for the same set of samples (e.g. insurance providers vs. research institutions). Transcriptomic data, gene expression data, can be horizontally shared across multiple hospitals or research institutions to increase patient cohorts and to increase genetic diversity for robust training of models. In any case, federated learning presents a range of both use cases and challenges, which we will outline in the subsequent paragraphs.\\

In classical FL, model parameters are sent to a central aggregator to form a global model, and this process is done iteratively until a suitable global model is established. However, the iterative nature of FL often leads to overhead and latency. Further, continuous communication could introduce an avenue of security threats such as data poisoning, model poisoning, backdoor attacks, and membership inference attacks \cite{mothukuri2021survey}. In contrast, \textit{one-shot federated learning} approaches such as \cite{guha2019one,hannemann2023privacy,hannemann2025private}, minimize this communication overhead by restricting communication to a single round. Each participant sends its relevant parameters to the central server once, which subsequently undertakes the role of model training. The central server computes the ML algorithm with a single round of communication. This approach drastically reduces the frequency of data exchange, which not only decreases the communication costs but also mitigates privacy risks associated with frequent updates. Furthermore, the central server can aggregate the received parameters efficiently, leading to fast model convergence. This makes one-shot federated learning particularly advantageous in scenarios with bandwidth constraints like hospitals without high-performance computing infrastructure. \\

FL however, is not fully a privacy-enhancing technology. FL does not provide any formal privacy guarantees, and it has been shown in multiple works, that FL is vulnerable to various privacy attacks \cite{boenisch2023curious,zhang2021survey}. However, FL can be seen as an initial step towards privacy-preserving ML, as it does not require the raw data to leave the client's device. In combination with differential privacy and/or secure multi party computation, FL can provide more robust privacy. 
\section{Related Work}
\label{sec:related}
This section reviews related work in the fields of FL and its applications in precision medicine, and discusses technical challenges and FL frameworks. \\

\paragraph{Precision Medicine using transcriptomic data.} With the ever increasing availability of large-scale transcriptomic data due to the advent of next-generation sequencing and single-cell measurements, the application of machine learning on transcriptomic data is a corner stone for today's precision medicine advancements. Based on this data, diagnosis can be improved, e.g. acute myeloid leukemia \cite{warnat2020scalable,eckardt2020application}, by the identification and validation of biomarkers and scores derived from machine learning. Additionally, typical fields of application are the cell and patient characterization like cell type annotation \cite{pasquini2021automated,yang2022scbert,chen2023transformer} or cancer subtype classification \cite{gao2019deepcc,khan2023deepgene}. Lastly, personalized medicine is powered by transcriptomic data and machine learning via the prediction of drug responses \cite{ji2021machine,qi2023trends} or individual survival probability \cite{gyHorffy2021survival,zhu2020application}. \\
In practice, cross-institutional pooling of transcriptomic data and relevant clinical meta-data information remains a challenge due to its sensitive character and large size. Even when limiting to non-raw data, it has been shown that genomic mutational signatures can be inferred from gene expression data \cite{oestreich2021privacy} highlighting the need for FL and privacy-preserving methods. In most real-world scenarios horizontal FL is and would be required to power machine learning on transcriptomic data since similar features (genes) are measured and heterogeneous patient cohorts (samples) can be combined in training for robust and generalizable models.
Further, cohort sizes often vary greatly and, lastly, hospitals often lacking high-performance computing power on-site, in particular GPU.   
Due to those reasons, federated learning is attractive solution for precision medicine applications.

\paragraph{Federated Learning and Applications}

FL is used for collaboration across medical institutions, hospitals, health care insurers or other entities \cite{zhou2022ppml,antunes2022federated,dayan2021federated,lee2020federated}. Some problems relevant to medicine were addressed by \cite{wu2022integrating} and \cite{danek2023federated} who trained a federated model to model Alzheimer's and Parkinson's disease. Beguier \textit{et al.} \cite{beguier2021differentially} introduce a differentially private and federated cancer occurrence prediction based on genomic data. Swaminathan \textit{et al.} \cite{swaminathan2024pp} present collaborative, multi-side and private genome-wide association studies on genomic data. For practical implications and benchmarking of frameworks for FL, however, there is less literature available. The multi-class data set we use in experiments \cite{hodge2019conserved}, to our knowledge, has not been modeled by any FL system. The authors of the binary data set propose a collaborative learning method named swarm learning without an aggregator \cite{warnat2021swarm}. This method achieves very good model quality, but does not take into account many challenges that arise when bringing FL in to production. 
We identified four major challenges with heterogeneity in data distribution, participating clients, consumption of computational resources and privacy:

\paragraph{Technical and Conceptual Challenges}

The issue of \textbf{statistical heterogeneity} of training data in FL encompasses both the distribution and size of the data. This challenge involves dealing with the non-IID (non-independent and identically distributed) nature of distributed data, which can lead to skewed or biased model training \cite{mendieta2022local,kairouz2021advances}. Additionally, the size of data sets of each client can vary significantly, where smaller data sets may not adequately represent the population, impacting model quality and generalizability. Fu \textit{et al.} \cite{fu2023client} showed that not only data heterogeneity can influence the model's quality, but also the varying number of clients which we call \textbf{architectural heterogeneity}. Navigating these aspects of statistical heterogeneity is crucial for ensuring the robustness and efficacy of FL models. In the context of transcriptomic data and health-care institutions both issues are very common since hospitals vary greatly in their sizes and specialization or different laboratories introduce bias due to small differences in sequencing protocols and machinery \cite{pfitzner2021federated}. 

Furthermore, while FL is designed to enhance \textbf{privacy} by training models locally and sharing only model updates, ensuring the privacy and security of these updates against potential inference attacks remains a critical concern. Multiple works showed, that there is no formal privacy guarantee for FL without additional privacy-enhancing techniques \cite{kairouz2021advances}. Recent publications \cite{boenisch2023curious,zhao2023secure} showed that baseline FL is vulnerable to reconstruction attacks, while others \cite{ganju2018property,melis2019exploiting} successfully performed Membership Inference Attacks (MIA). Multiple works \cite{choudhury2019differential,beguier2021differentially,adnan2022federated} explored Differential Privacy in a FL scenario for medical data. Differential Privacy protects the privacy of individual data points in a training data set while allowing ML models to benefit from the overall information. It adds controlled noise to data or model parameters, making it difficult to infer specifics about individual entries. However, the application of noise usually comes with information loss. This can pose challenges for real-life applications \cite{hannemann2024differentially}. For transcriptomic data in combination with clinical patient data, this is a dilemma since biomarker signals are often weak for multi-factorial diseases. Hence, privacy levels need to be carefully chosen to find a trade-off between model quality, which is highly critical in medical applications, and privacy. 

Finally, \textbf{resource consumption} presents another challenge for FL. The diverse and potentially resource-limited nature of participating clients in a FL network can lead to inefficiency and delay \cite{kairouz2021advances,fu2023client}. Furthermore, the communication required for model updates and synchronization in FL adds to network bandwidth demands, which can be a bottleneck in resource-limited environments. 

\paragraph{FL Frameworks}
The field of federated learning is rapdily evolving, and there are many existing open-source FL frameworks, such as TensorFlow Federated (TFF) \cite{abadi2016tensorflow,tf}, Flower (FLWR) \cite{beutel2020flower}, PySyft \cite{ziller2021pysyft}, FATE \cite{liu2021fate} and FedML \cite{he2020fedml}. 
These frameworks vary in terms of their features, ease of use, and specific use cases. The choice of a federated learning framework typically depends on the specific requirements and constraints of the application.

Both PySyft and TFF are well established and benefit from a large community support. While TFF is based on the TensorFlow ecosystem, PySyft works primarily with PyTorch. Both PySyft and FATE provide multiple optional privacy-enhancing methods such as Differential Privacy and Secure Multi Party Computation. FLWR is designed to be framework-agnostic and can work with various machine learning frameworks, including TensorFlow, PyTorch, and others. In terms of abstraction level, Flower's API is more high-level and is, therefore, supposed to be more user friendly than TFF and PySyft. While FLWR and FATE only allow simulation and cluster deployment, FedML provides a flexible and generic API and allows on-device training. Also, FedML can be used for various network topologies such as Split Learning, Meta FL and Transfer-Learning.  \\

For FL in the biomedical domain several specialized platforms and frameworks have been developed. These platforms and frameworks such as FeatureCloud \cite{matschinske2023featurecloud} or sfkit \cite{mendelsohn2023sfkit} combine a special focus on privacy-preserving FL and user-friendly interfaces for interdisciplinary research. Further, FeatureCloud is unique in its app-like platform and community-based approach for the development of functionalities. Lastly, swarm-learning \cite{warnat2021swarm}, a specialized FL approach, has been developed on genetic data for secure cross-institutional FL. \\

\section{Methodology}
\label{sec:method}
This section explains our experimental concept, the data sets we used, and the architectures of the machine learning models for our experiments. 

\subsection{Experimental Concept}
To quantify the impact of our four key challenges on FL with transcriptomic data, 
we measure the quality of a global model obtained by FL on centralized data first. With this baseline, we conduct comparative experiments to explore the effects of \textit{Architectural} and \textit{Statistical Heterogeneity} 
on the model quality. We explore the impact of \textit{Gaussian noise} to enhance privacy and measure its effect on the model. Furthermore, we assess the \textit{Resource Consumption} at the aggregator and the clients. 

To draw robust conclusions, we vary problem type and model architecture. In particular, we conduct experiments not only on a binary-labeled data set but also on a multi-class data set. This aligns with clinical research, which typically covers a variety of diseases and research questions rather than a single condition. In multi-class problems, the complexity increases as the model must differentiate between multiple, often overlapping conditions.


\subsection{Data Sets}
We use two  data sets, whose statistical details can be found in \ref{tab:data-statistics}.
The Acute Myeloid Leukemia data set~\cite{warnat2020scalable} was previously obtained from 105  studies, resulting in 12,029 samples with binary labels. 
We call it the \textbf{binary-labeled data set}. It consists of gene expressions by microarray and RNA-Seq technologies from peripheral blood mononuclear cells (PBMC) of patients with either a healthy condition or acute myeloid leukemia (AML). 

The \textbf{multi-class data set} includes expression profiles generated by single-cell RNA-Seq for cell types of the human brain, in particular the middle temporal gyrus (MTG). The data set was published by \cite{hodge2019conserved}, who isolated sample nuclei from eight donors and generated gene expression profiles by single-cell RNA-Seq for a total of 15928 cells (samples) describing 75 distinct cell subtypes. We reduced the number of classes (cell types) to make the data set more suitable to experiment with class imbalance. For that we selected only the the five most abundant cell types (classes) leading to 6931 cells (samples) for training. We preprocess both data sets as in previous analyses and benchmarks \cite{warnat2020scalable,warnat2021swarm,hodson2016precision,hodge2019conserved}. 

\begin{table}[!ht]
    \caption{Statistics of Data Sets Used in the Experiment}
    \label{tab:data-statistics}
    \begin{center}
    \resizebox{\textwidth}{!}{%
    \begin{tabular}{lcccc}
    \toprule
    Data Set & Data Points & Dimensions & Label Type & Distribution \\
    \midrule
    Acute Myeloid Leukemia  & 12,029 & 12,708 genes & Binary & Imbalanced \\
    Middle Temporal Gyrus    & 15,928 & 13,945 genes & Multi-Class & Imbalanced \\
    \bottomrule
    \end{tabular}
    }
    \end{center}
\end{table}

\subsection{Model Architectures}
We experiment with a \textbf{logistic regression model }and a \textbf{deep-learning model}. 
Following~\cite{warnat2021swarm}, the deep-learning model uses a sequential neural network architecture. It consists of a series of dense layers, each with 256, 512, 128, 64, and 32 units, all activated using the 'relu' activation function. Dropout layers with dropout rates of 0.4 and 0.15 
prevent overfitting. The configuration of the output layer is based on the number of classes in the data set. 

\begin{table}[ht]
\centering
\caption{Optimized Hyperparameters}
\label{tab:model-parameters}
\resizebox{\textwidth}{!}{%
\begin{footnotesize} 
\begin{tabularx}{\linewidth}{XXX} 
\toprule
\textbf{Data Set} & \textbf{Model} & \textbf{Hyperparameters} \\
\midrule
\multirow{2}{*}{Binary} & Deep Learning & Adam, L2: 0.005, Epochs: 70 \\
\cmidrule{2-3}
& Logistic Regression & SGD, L2: 0.001, Epochs: 8 \\
\midrule
\multirow{2}{*}{Multi-Class} & Deep Learning & Adam, L2: 0.005, Epochs: 30 \\
\cmidrule{2-3}
& Logistic Regression & SGD, L2: 1.0, Epochs: 10 \\
\bottomrule
\end{tabularx}
\end{footnotesize} 
}
\end{table}

For our baseline and for hyperparameter tuning, we train both models on centralized data. 
In particular, the hyperparameter space was randomly searched to find Cross Entropy as optimal loss function with a batch size of 512. Hyperparameters that differ in the respective combinations of model and data are summarized in Table~\ref{tab:model-parameters}. 
While the parameters were optimized in a centralized learning setting, we applied them for consistency to our federated models.

We denote the rounds of training based on the local epochs, so that the total number of epochs remains a constant. Assume one round of training and two clients using 100 local training epochs. With two rounds of training, this would be 50 local epochs for both clients. 

\subsection{Frameworks}
The choice of the frameworks for our study was driven by multiple factors. First, we prioritized documentation and usability. Additionally, our interest in exploring both horizontal FL and evaluating frameworks with varying levels of abstraction led us to test one with high abstraction and another with low abstraction. Finally, the same communication protocol was relevant for our experiments. This selection process aimed to assess and compare TensorFlow Federated and Flower \cite{abadi2016tensorflow,tf,beutel2020flower}.

\section{Experiments}
\label{sec:experiments}
In this section, we describe our experimental setup and our analysis results regarding model quality, data quality and resource consumption. 

\subsection{Experimental Setup}
We perform all experiments using one CPU core from an AMD(R) EPYC-(R) 7551P@ 2.0GHz - Turbo 3.0GHz processor and 31 Gigabyte RAM for each client. The network is a 100 Gbit/s Infiniband. We measure the network traffic with tshark \cite{tshark}. No GPU is used during the experiments. To ensure resource parity among different frameworks and the central model, each training process is bound exclusively to one CPU core. Our experiments are implemented in Python. 
For preprocessing and data loading, we used the libraries Pandas \cite{mckinney2010data} and Scikit-learn \cite{pedregosa2011scikit}. 
Both the logistic regression model and the deep-learning model were implemented using Keras, with default settings and federated algorithms from FLWR and TFF~\cite{abadi2016tensorflow,tf,beutel2020flower}. 
The default of FLWR is a federated averaging strategy in a client-aggregator setup. Additionally, we 
compute the average score of each metric for every client. The default building function in  TFF uses a robust aggregator without zeroing and clipping of values as model aggregator. The clients of TFF all use the eager executor of TFF and are loading the data with a custom implementation of the data interface of TFF \cite{abadi2016tensorflow}. In this configuration, FLWR and TFF implement the federated averaging algorithm with a learning rate set to 1.0~\cite{mcmahan2017communication}. The code to our experiments can be found at~\cite{leoscode}.

For our experiments,  
we tested combinations of Logistic Regression (LogReg) and Sequential Deep Learning (DL) models together with binary problems (Binary) and multi-label problems (Multi). In the figures, we abbreviate the combinations of models and problem types as Binary LogReg, Multi LogReg, Binary DL and Multi DL. For each combination, we tested 3, 5, 10, 50 clients and 1, 2, 5, 10 rounds of training, and we measured model quality and computational resources used. To analyze the effect under investigation, we iterated over the respective other parameter and reported the averaged result, i.e., when varying the number of clients, we conducted tests for each training round configuration and presented the cumulative effect observed across all training rounds.

\subsection{Model Quality}
To explore the impact of heterogeneity on the global model, we use a 5-fold cross validation and compute the Area Under the Curve (AUC). A higher AUC value (closer to 1) indicates a better model, that distinguishes between the classes more effectively. 
We compare the AUC of the centralized baseline with the AUC obtained with FL and varying numbers of clients and training rounds.
%
%
In the following, we explain our key findings. 

\newcounter{findingscnt}
\newcommand{\myfinding}[1]{\stepcounter{findingscnt}\textbf{\newline Finding \thefindingscnt: #1}}

\begin{figure}[hb]
    \centering
        \vskip -0.1in
    \includegraphics[width=1\linewidth, trim={2 2 2 2}, clip]{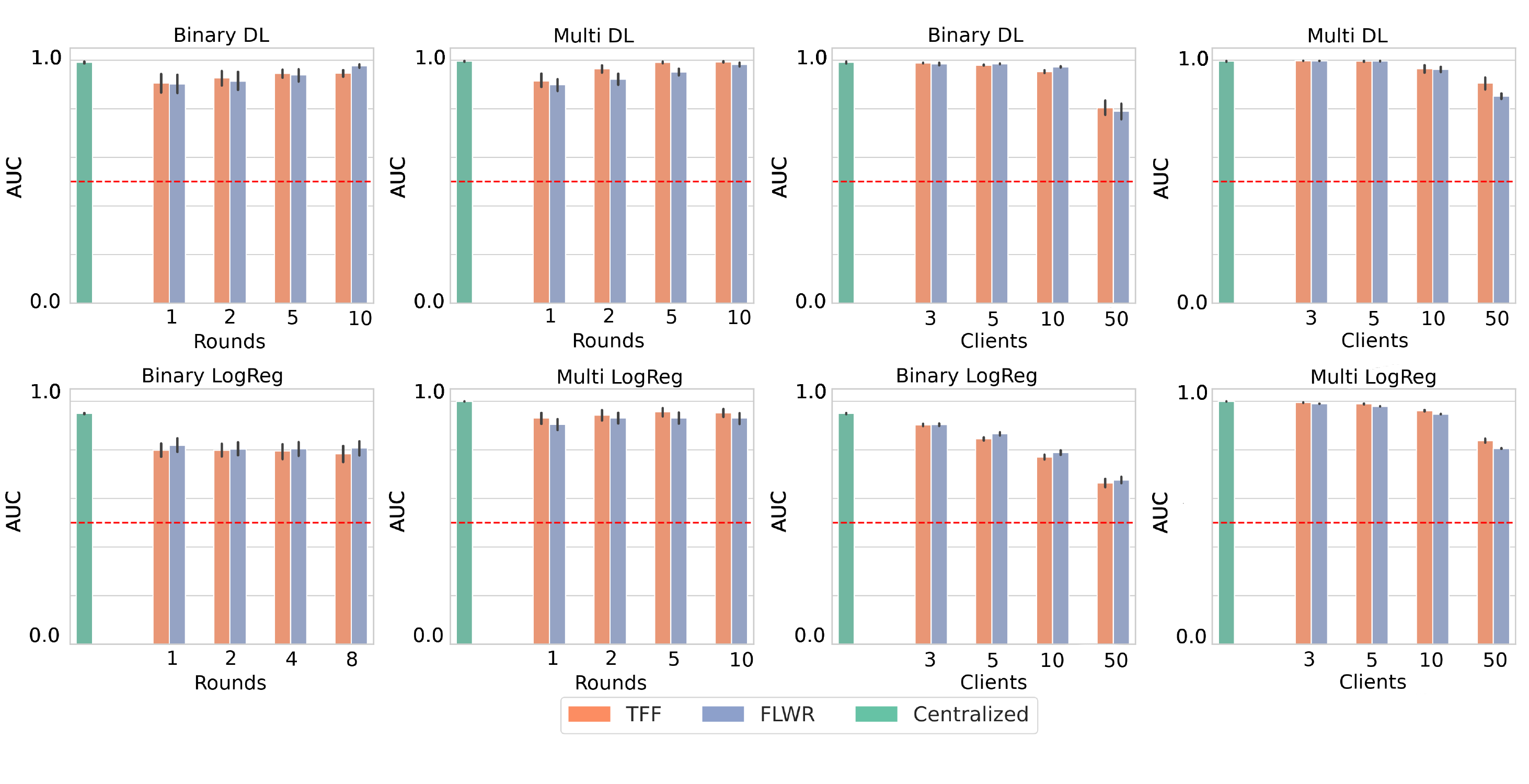}
    \vskip -0.1in
    \caption{AUC with respect to an increasing training rounds and clients}  
    \label{fig:balanced}
    \vskip -0.1in
\end{figure}

\myfinding{Boosting training rounds does not always enhance model quality.}
We assumed from previous work (cf. Sec. \ref{sec:related}) that an increasing number of training rounds improves the quality of the global model.
To examine the impact of frequency of weight updates among clients during training, we varied the numbers of training rounds and kept the total number of epochs constant. This approach is consistent with our round configurations optimized with hyperparameter tuning, as described in Section \ref{sec:method}.
Consider Figure \ref{fig:balanced}. The left two columns of diagrams show the results of our experiments with varying numbers of rounds, the right ones varying numbers of clients over the respective other variable. The top line of diagrams were obtained with deep learning models, the bottom line with logistic regression. We find that the AUC of the global model does not improve significantly with the number of rounds for logistic regression. Deep learning benefits slightly with more rounds of weight updates (see Figure \ref{fig:balanced}).

\begin{figure}[!hb]
    \vskip -0.1in
    \centering
    \includegraphics[width=1\linewidth, trim={2 4 4 2}, clip]{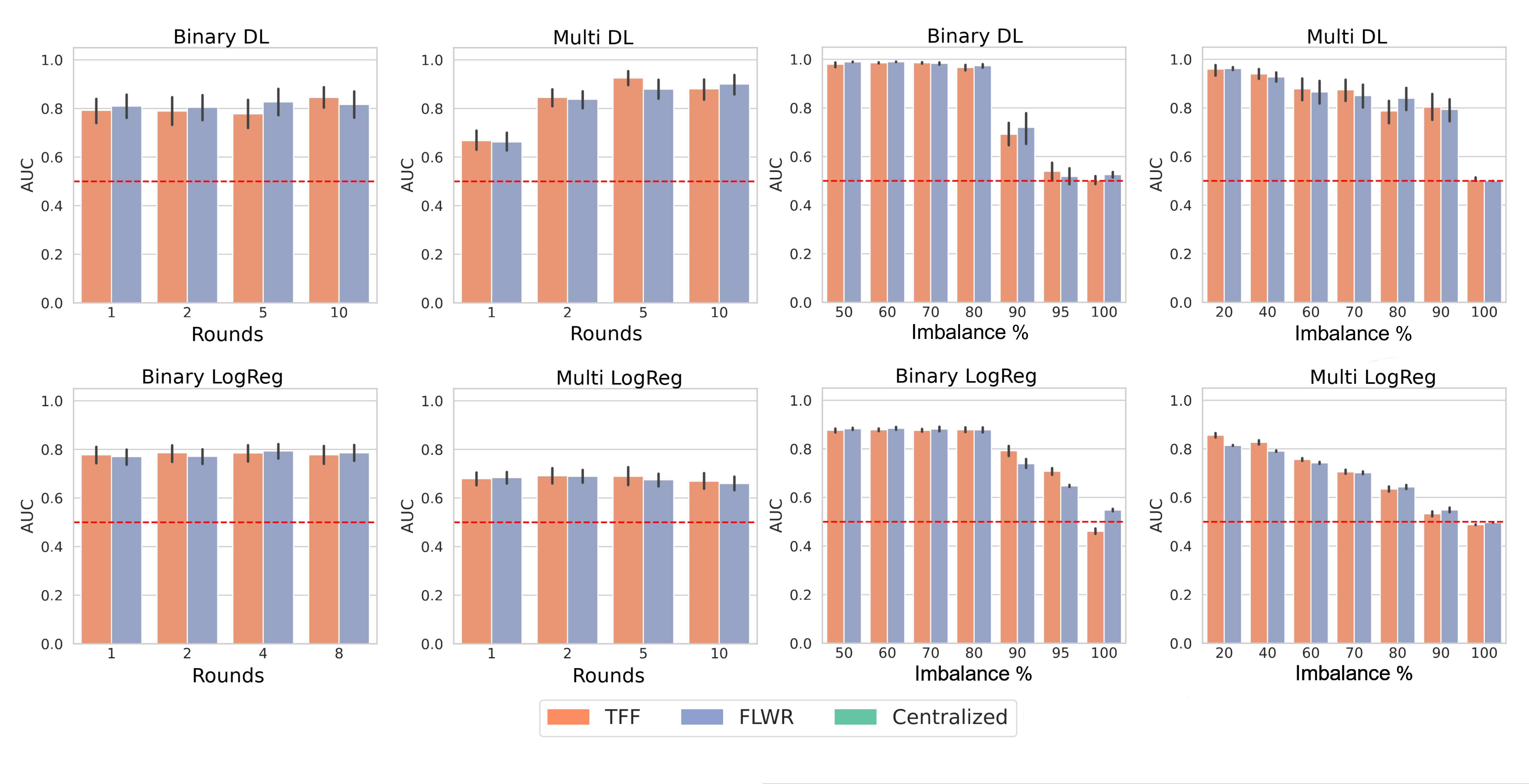}
    \vskip -0.1in
    \caption{AUC with respect to an increasing class imbalance and training rounds over all imbalance configurations}
    \label{fig:imbalanced_DL}
\end{figure}

For non-balanced data sets, matching most real-world FL scenarios, updating rounds prove to be more effective to mitigate class-imbalance across different clients (see Figure \ref{fig:imbalanced_DL}). Thus, healthcare institutions should carefully chose the number of rounds in FL applications, based on the data distribution and machine learning algorithms used.

\myfinding{Fewer clients and more data increases model quality.}
To analyze the effect of the number of clients, we split the training data in disjoint subsets and distributed them to clients. Each subset has the same size and class distribution as the whole data set. 
The data on each client is then split into training and test data with a ratio of 80:20. The test data of all clients is combined and the aggregated FL model is evaluated on this test data. 
Then, both FL frameworks were used to train a global model. 
For a small number of clients, Figure \ref{fig:balanced} reports that the AUC of FL is similar to our baseline (AUC 0.98),  proving that FL can reach centralized model quality in real-world scenarios. A slight decrease for 10 clients can be discovered which is followed by a drop in AUC in the extreme case of 50 clients. This results from the reduced size of local training data: As the number of clients increases, the data set is divided among them, resulting in a reduction in the training data available for local training. 
The limited data size does indeed reflect a possible scenario where small healthcare institutions want to attend to a FL scenario. 
We conclude that clients should only allowed
when they provide a substantial number of samples on which local model training can be performed. 

\myfinding{Model quality is driven by models, not by frameworks.}
Figure \ref{fig:balanced} shows that the logistic regression model has an overall lower model quality than the deep learning model for the multi-class problem, as its highest AUC value is 0.90 for both frameworks. In contrast, the deep learning model has an AUC of 0.99 for TFF and 0.98 for FLWR.
This emphasizes the superiority of  deep learning for this data set, and underlines that the model (and their hyperparameterisation) must fit to the problem. Benchmarks for a direct comparison of model quality are rare, especially for transcriptomic data. We wanted to learn if there is a difference between FLWR and TFF across the tested scenarios. As Figure \ref{fig:balanced} illustrates, if all parameters and aggregation algorithms for FLRW and TFF are configured in the same way, both frameworks deliver a similar model quality. This observation holds for various configuration. 
Healthcare institutions are therefore free to select FL frameworks based on functionality (e.g. privacy-support), usability and computational resource demand, instead of concerning model quality. 

\myfinding{Class imbalance impairs federated learning.}
A challenge for FL is that data points are usually not independent and identically distributed (IID), leading to statistical heterogeneity. With transcriptomic data, a balanced class distribution among all clients seems unlikely. Therefore, we explored the effect of data imbalance on the global model's quality. We investigate IID- and non-IID-data by methodically increasing the imbalance to find a sweet spot of imbalance and model quality.
We start our experiments with a number of clients equal to the number of classes, and all classes are equally distributed among the clients. Subsequent experiments increase the number of samples from one class while reducing the number of samples from another class. This process is repeated independently for each client and class, until each client contains samples from a single class. We present a visualization of the conducted experiments in Figure \ref{fig:imbalanced_DL}. 
The evaluation of model quality in these scenarios is conducted using a separate test data set characterized by an equal distribution of classes.
We compare the resulting AUC with an equally-distributed data set. As Figure \ref{fig:imbalanced_DL} shows, deep learning can indeed fight class imbalance. However, if the imbalance exceeds a certain threshold, a sudden drop in AUC can be expected.  

The threshold depends on the data set and machine learning algorithm. Our results indicate that a non-IID distribution not necessarily results in poor model quality. But, the model quality in the presence of non-IID is strongly dependent on the problem type and data set. This can be further increased with the appropriate model selection. As Figure \ref{fig:imbalanced_DL} shows, deep learning is robust with a drop in AUC at 90\% imbalance. 
Whereas more training rounds does not improve the robustness for logistic regression, it does for deep learning. We also investigated the effect of multiple rounds to a non-IID setting. Again, we increased the training rounds over all configurations in data distributions and report the average. 
The increase in a number of training rounds leads to an improvement of model quality for deep learning with non-IID data, but does not affect the quality of logistic regression, regardless of the problem type (see Figure \ref{fig:imbalanced_DL}). 
Thus, healthcare institutions need to consider that the model quality depends on minimal number of samples per class. This number depends on aspects like the training algorithm.
Based on these findings medical institutions need to consider that, while FL is mostly robust to class imbalance, a minimal number of samples per class is necessary and that this minimum is highly dependent on other aspects like the machine learning algorithm used for training.

\subsection{Data Quality and Privacy}
There are many anonymization approaches such as Differential Privacy, which apply noise to the data. Because TFF and FLWR have different levels of support for Differential Privacy, we have chosen a general approach to investigate the impact of anonymization on model quality: 
We add Gaussian noise to the local parameters before aggregation. 

\begin{figure*}[h!]
    \vskip -0.1in
    \centering
    \includegraphics[width=1\linewidth, trim={5 5 5 5}, clip]{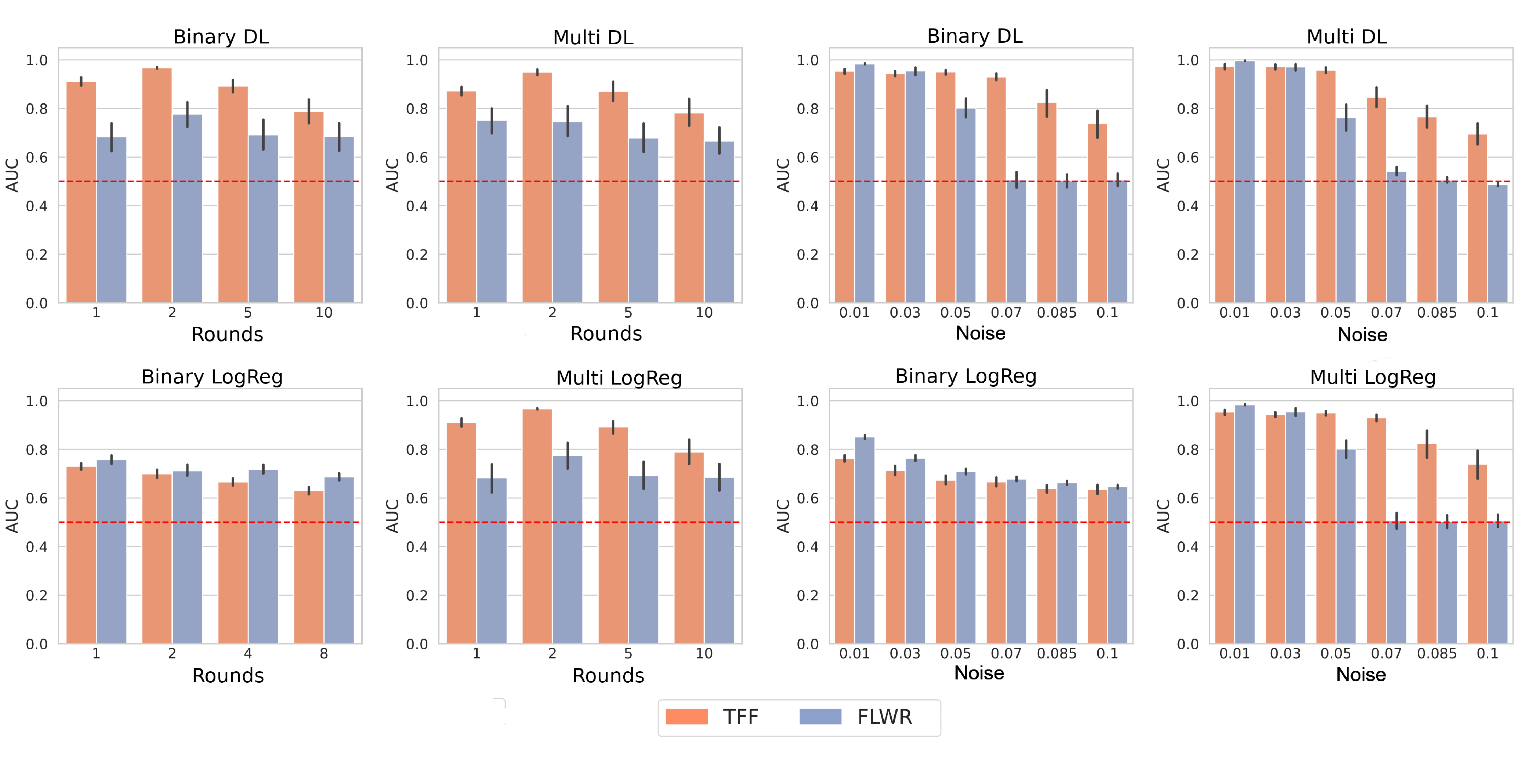}
    \vskip -0.1in
    \caption{AUC with respect to an increasing local Gaussian noise}  
    \label{fig:AUC}
\end{figure*}

\myfinding{Adding noise to the weights has a strong impact on model quality.}
Our experiments use five clients and varying noise parameters 0.01, 0.03, 0.05, 0.07, 0.085, 0.1.  
We observe that all model and problem types can deal with some noise. However, at some point AUC drops to approx. 0.07 -- 0.085. TFF copes with noise better than FLWR, possibly because TFF uses regularization techniques that mitigate the impact of noisy updates. 
Increasing the training rounds does not improve the model quality, but slightly decreases AUC. This is due to the noise being added before aggregation. As shown in Figure \ref{fig:AUC}, the greater robustness of TFF to noise is evident once again.
This investigation of Privacy-enhancing Gaussian noise provides implications and guidelines for FL in medical scenarios. Firstly, FL does not guarantee privacy and, depending on method, privacy-level must be carefully chosen to ensure model performance. Secondly, FL frameworks show differences both in supported functionality to implement privacy, as well as global model performance.

\subsection{Computational Resources}
When analyzing the computational resources of a federated system, both the local and the global perspective are relevant. Here, the local perspective refers to the resource consumption of each individual client, while the global
\begin{figure*}[!ht]
    \centering
    \includegraphics[width=1\linewidth, trim={5 5 5 10}, clip]{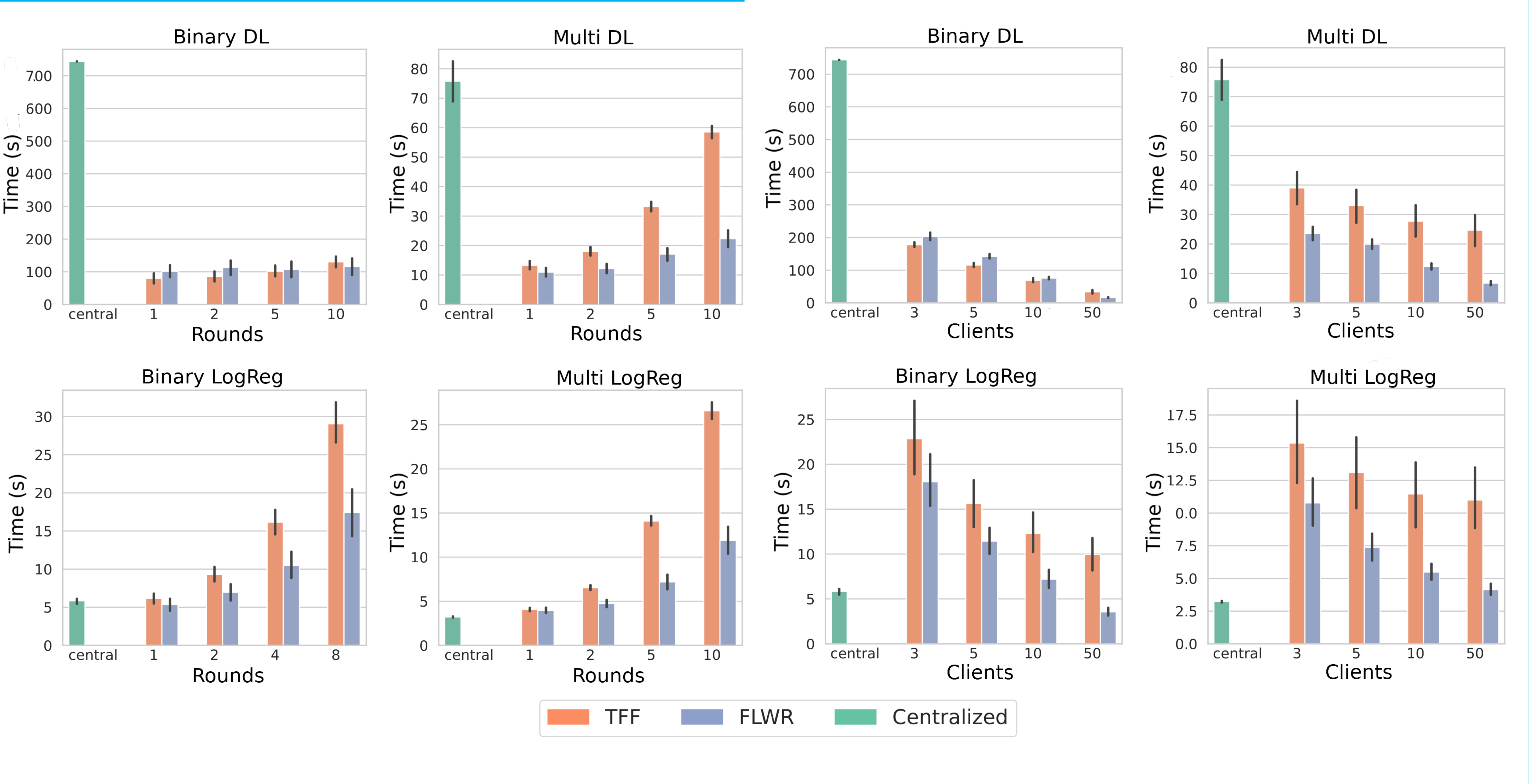}
    \vskip -0.1in
    \caption{Global training time with respect to increasing training rounds and clients}  
    \label{fig:time2}
\end{figure*}
perspective refers to the total consumption of the entire system, including all clients and the aggregator. The local resource consumption is driven by the common constraint of limited client resources, making it relevant for any real-life application. In contrast, global resource consumption becomes relevant when the goal of the federation is to parallelize training and, therefore, optimize the total training time.
We investigate the training time, memory consumption and network traffic both locally and globally. 

\begin{figure}[!h]
    \centering
    \includegraphics[width=0.6\linewidth, trim={2 2 2 2}, clip]{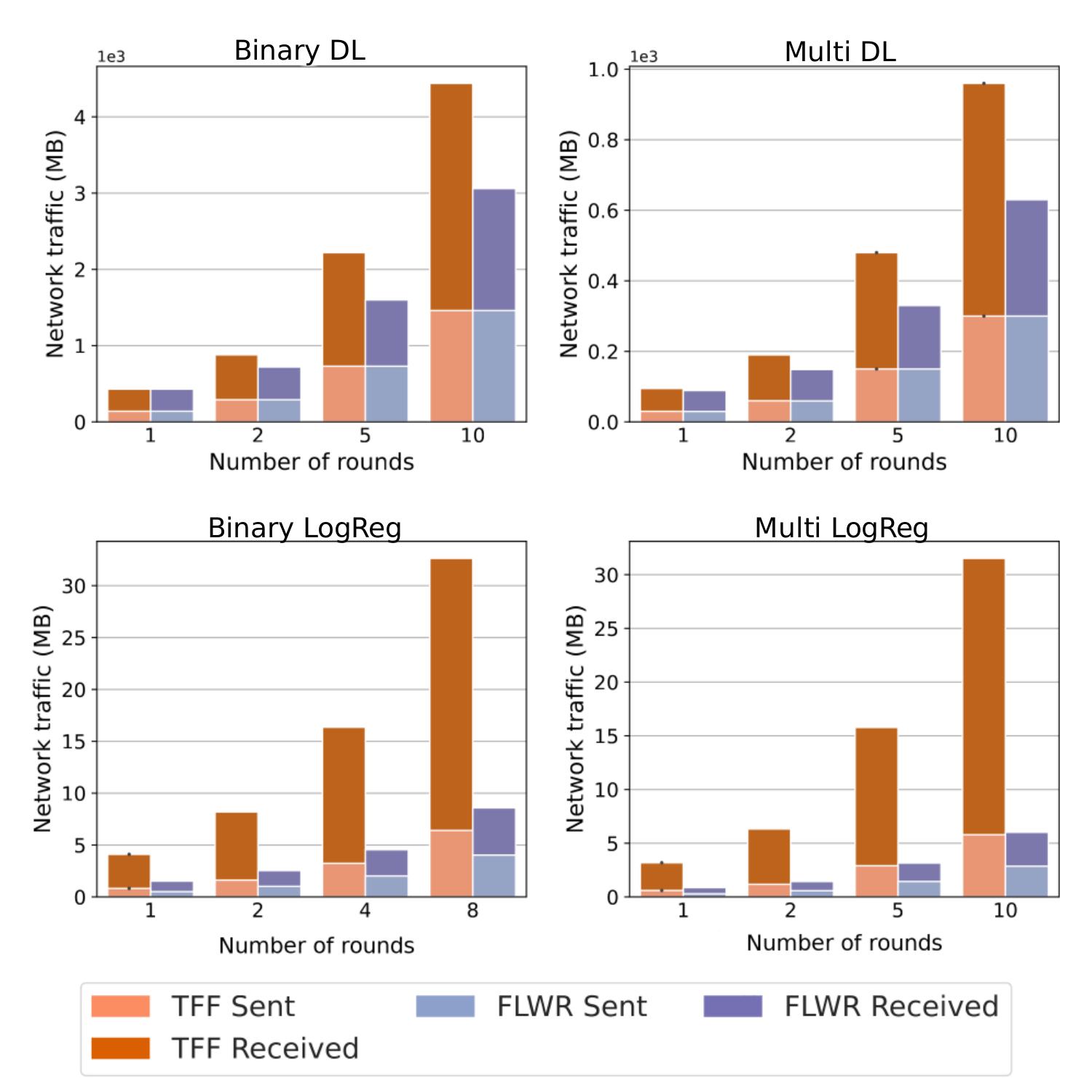}    
    \caption{Network traffic with respect to an increasing number of rounds}  
    \label{fig:network_traffic}
\end{figure}

\myfinding{Global training time is highly dependent on various factors.}
When training on highly dynamic data, global training becomes particularly important. In such cases, the system must efficiently handle new data and frequent updates  across clients, which impacts overall training time and, thus, decision making. Here, we analyzed the global training time by measuring the time required to train local models, transmit model parameters, aggregate them, and return them to the clients. Again, this was tested over multiple rounds and with various clients; the results are reported in Figure \ref{fig:time2}.
Federated training of deep learning models generally outpaces centralized training due to parallelization across multiple clients. However, for logistic regression, centralized training tends to be faster, as the simplicity of the model reduces the benefits of parallelization in federated settings. Interestingly, the implementation using FLWR was up to twice as fast than that of TFF. Additionally, the increased network traffic (see Figure \ref{fig:network_traffic}) might slightly increase the total training times.
\begin{figure*}[!htb]
    \centering
    \includegraphics[width=1\linewidth, trim={10 5 5 10}, clip]{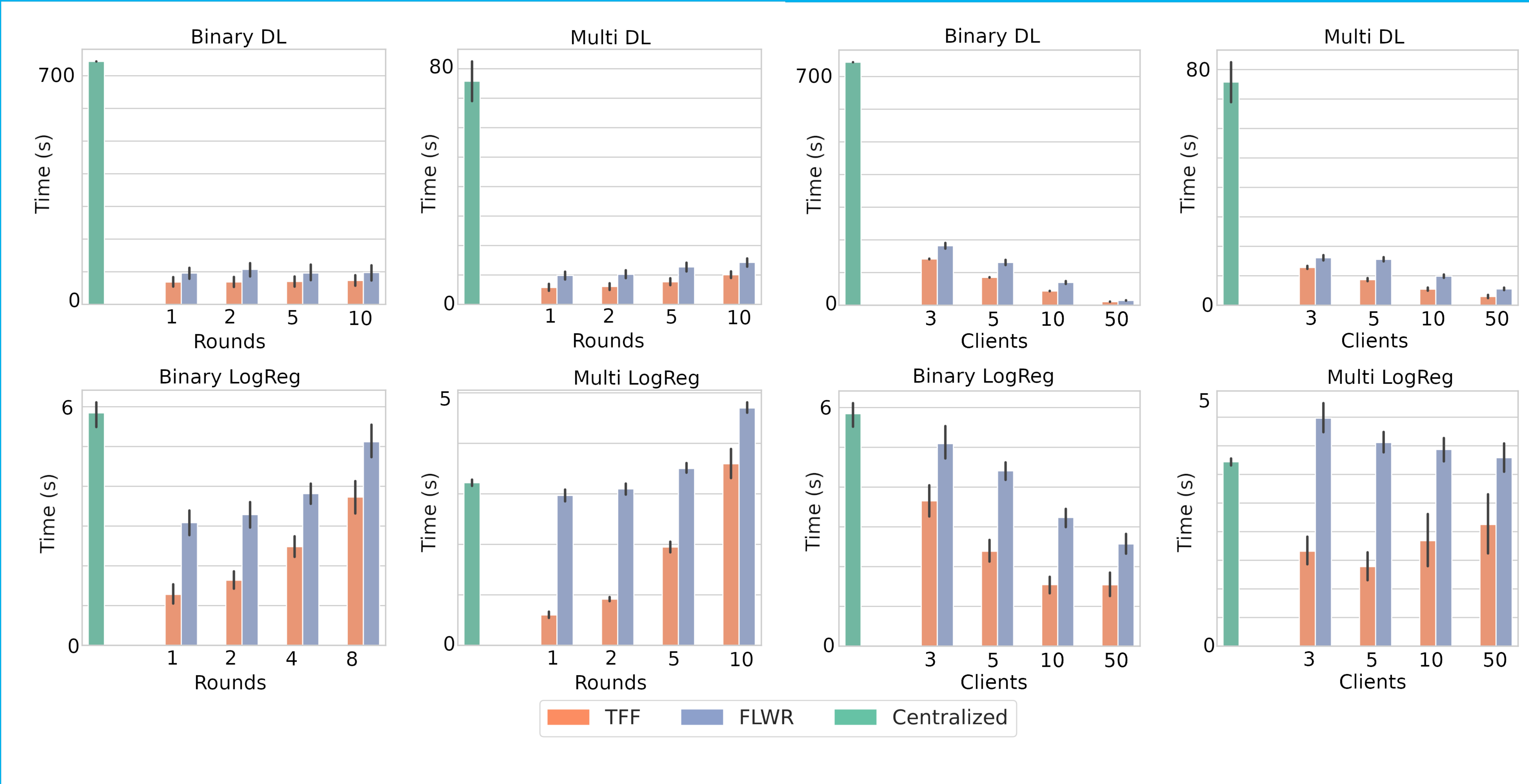}
    \vskip -0.1in
    \caption{Local training time with respect to increasing training rounds and clients}  
    \label{fig:time}
    \vskip -0.1in
\end{figure*}

This depends on the number of rounds and clients, and the network capacity and latency of the coordinator. 
In conclusion, if the overall training time of the federated system is critical for the healthcare application, careful model selection is essential. The use of FLWR could be beneficial as well.

\myfinding{FL does not increase local training times.}
We measure the training time for each client individually over multiple training rounds. 
Recall that we keep the total number of training epochs and the total number of samples constant, i.e., the more clients, the fewer local training rounds and the smaller the sample sizes per individual client. Thus, we assume that more clients result in smaller training times per client. Figure \ref{fig:time} confirms this.
The figure is organized in the same way as Figure \ref{fig:balanced}, i.e., models in rows and problem types in columns. 
TFF provides a much faster local training compared to FLWR, because of differences in their implementation. 
The difference between centralized training and federated training is less distinct for logistic regression than for training deep learning models. 
Thus, healthcare institutions benefit more from federated training for complex machine learning approaches with long training times such as deep learning models.

\begin{figure*}[ht]
    \centering
    \includegraphics[width=1\linewidth, trim={5 5 5 10}, clip]{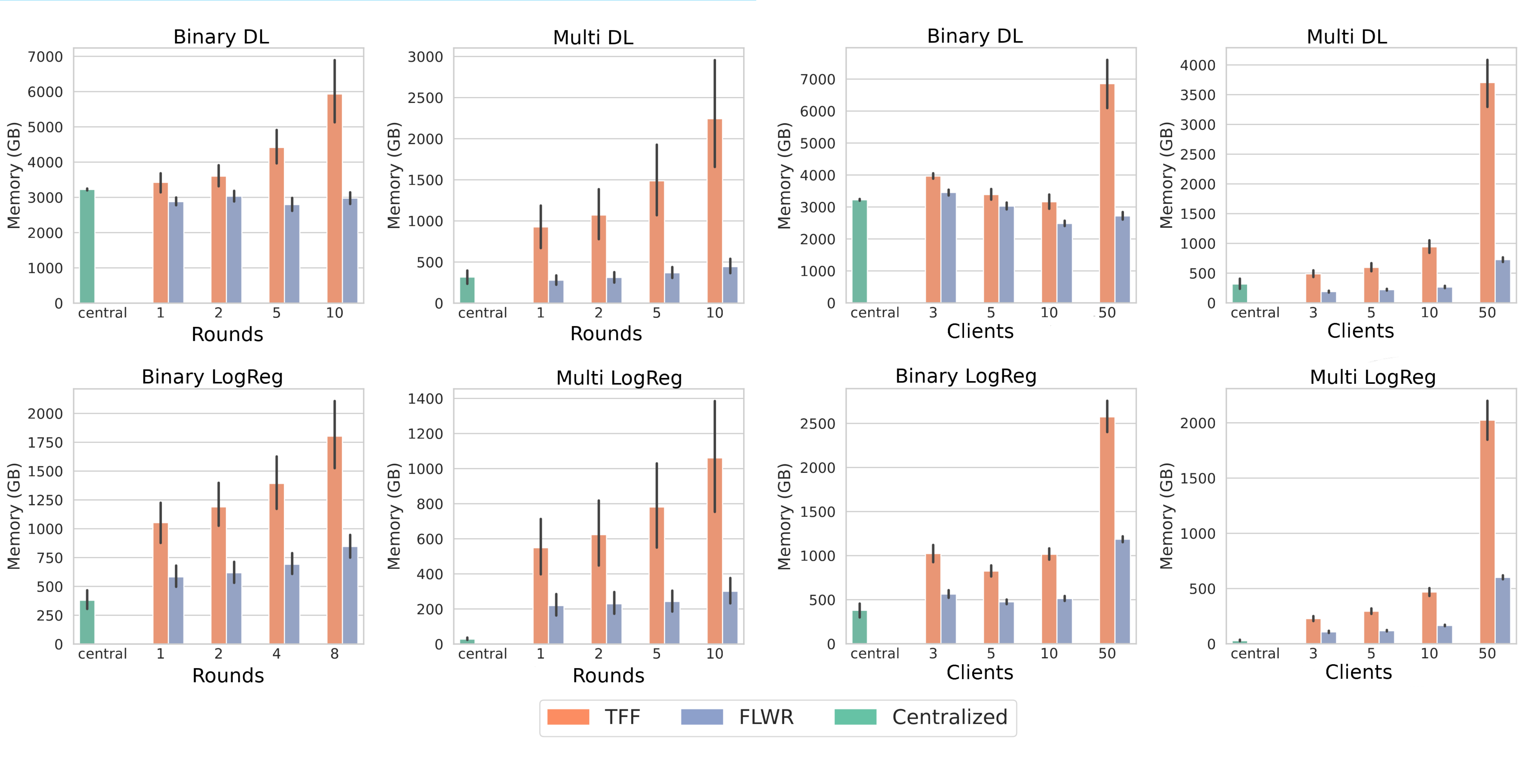}
    \vskip -0.1in
    \caption{Global memory consumption with respect to increasing training rounds and clients}  
    \label{fig:memory2}
\end{figure*}

\begin{figure*}[!h]
    \centering
    \includegraphics[width=1\linewidth, trim={5 5 5 10}, clip]{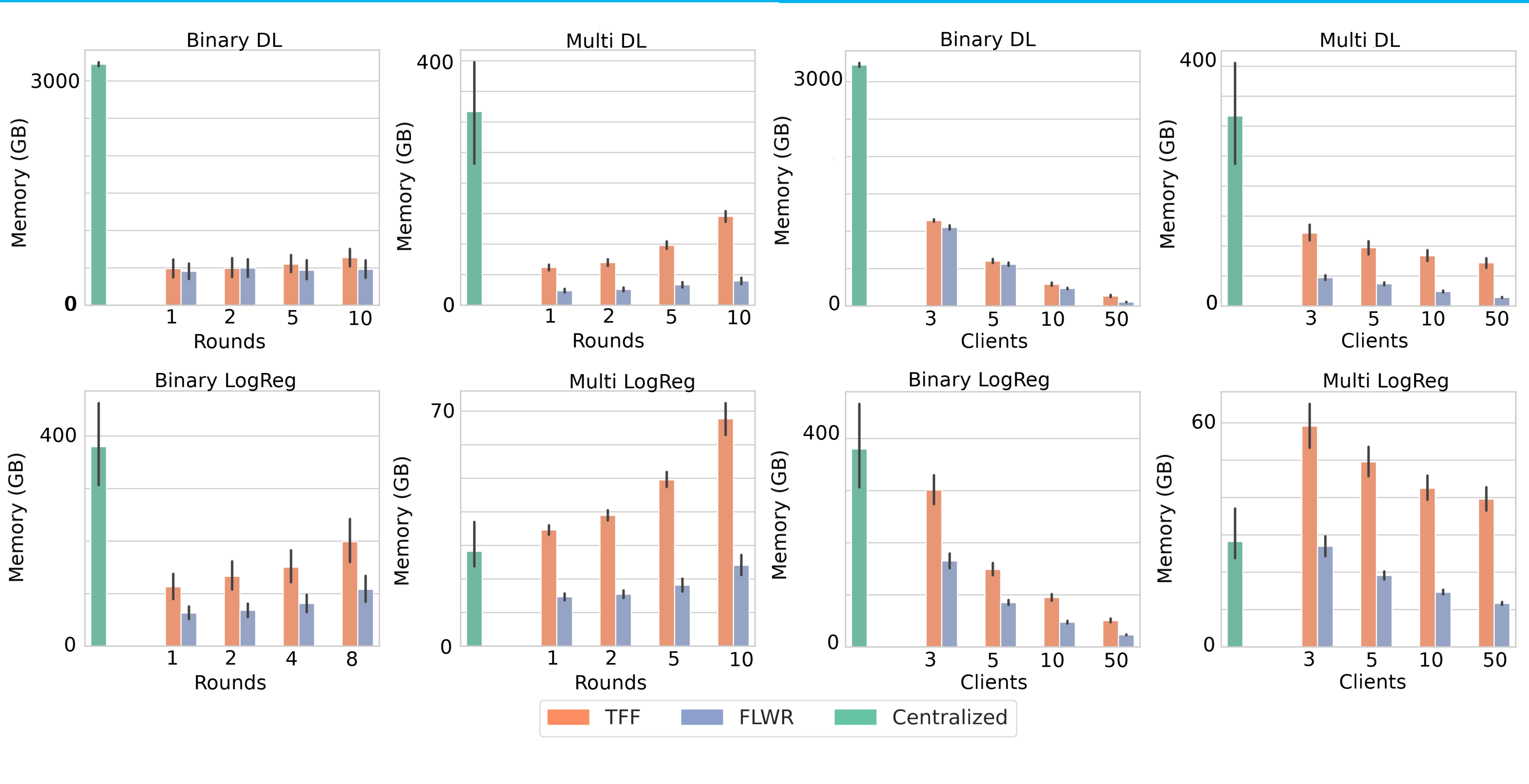}
    \vskip -0.1in
    \caption{Local memory usage with respect to increasing training rounds and clients}  
    \label{fig:memory1}
\end{figure*}

\myfinding{FL does increase the global memory consumption.}
In Figure \ref{fig:memory2}, the global memory consumption over increasing training rounds and clients is reported. The global memory consumption of the federated system is considerably greater than that of centralized computation. This is consistent across both model types and problem types. The increased memory usage in federated systems stems from the need to store multiple local models, intermediate parameters, and aggregated results from all participating clients. This seems to be critical, especially when healthcare institutions have limited resources. However, this overhead is distributed among clients, resulting in limited overhead for each individual client, as discussed in the following finding.

\myfinding{Memory consumption is effectively shared.}
We measured the aggregated memory consumption for both the clients and the overall system. 
As the number of clients or rounds increases, the overall resource consumption increases due to the increase in coordination effort. However, the client-wise resource consumption decreases, which is an advantage for healthcare institutions. 
We observed that both the clients and the global FL system as a whole require more memory with deep learning than with logistic regression.
This was expected, because deep learning uses much more parameters than logistic regression. Second, TFF uses much more memory than FLWR (Figure \ref{fig:memory1}, again with models in rows and problem types in columns).
We conclude that FL saves healthcare institutions a significant amount of memory, at the cost of a slight increase in global training time and global memory requirements. 
Further, FL frameworks show distinct differences in their training times and memory consumption revealing potential for optimization of FL tools.

\myfinding{The network load is not a bottleneck.}
To assess the network load, we assume that the amount of data transmitted and received is determined by the data serialization method of the framework, and is not influenced by hardware or interference from other clients. Therefore,  we experiment with a fixed number of 10 clients. In accordance to the increased memory usage of TFF, TFF comes with higher network traffic as well. Additionally, since deep learning needs to share more parameters than logistic regression, the network traffic rises from 4 MB (peak for LogReg) up to 30 MB (peak for DL). 
For comparison, an average household has a bandwidth of 209 Mbps and can easily handle the network demands \cite{fairinternetreport}. We conclude that the network traffic is acceptable for healthcare institutions, and may pose a problem only for the training of very large neural networks such as foundation models. The frameworks have different demands on computing resources, but this is not a basis for selection in most medical scenarios, and it does not restrict the applicability of FL on transcriptomic data.

\section{Discussion} \label{sec:discussion}
Our analysis highlights several key findings, both challenges and benefits, of applying FL to transcriptomic data. With regards to model quality, several effects were evident: 
Architectural and statistical heterogeneity remain central concerns, as variations in client infrastructure and data distributions affect model performance. For multi-factorial problems and high number of features, deep learning models outperform logistic regression models in terms of model quality. Increasing the number of training rounds does not greatly improve the global model quality, showing the high effectiveness of weigh aggregation in FL. Nevertheless, hyperparameter tuning has a large impact, often outweighing the effect of increasing training rounds. This suggests that effective parameter selection and optimization strategies are more critical than merely extending training. Transcriptomic data is robust to some class imbalance, especially using deep learning models. Problem type and data set are key factors for robustness, also with respect to the amount of training data. Finally, privacy-preserving Gaussian noise can lead to a drastic loss in model quality, underlining the trade-off between privacy and model quality. \\

With regards to efficiency, our experiments present valuable practical insights: FL saves memory and training time for individual clients. The more clients, the lower the individual load. Interestingly, Flower consumes less computational resources than TensorFlow Federated, additionally requires less domain knowledge about FL, but is also less customizable. Despite additional network traffic, our measurements indicate that FL is feasible for typical health institutions and their existing compute infrastructure.\\

Overall, FL presents a practical alternative to centralized learning for common tasks in precision medicine such as disease prognosis and cell type classification. However, future research in this field is warranted in multiple directions: First, privacy-preserving techniques should be explored further. In this work, we assume trusted parties. This can be reasonable if the clients grant access to the data upon approval of an application, which is current practice in biomedical research, but creates additional bureaucracy and time delays. Data collaboration between different data holders, without excessive access application processes, may require the implementation of additional privacy-enhancing mechanisms, as federated learning does not fully protect against privacy leakage \cite{boenisch2023curious}. Second, to make the findings more applicable, future work could focus on enhancing robustness and generalizability of models in precision medicine. Here, more advanced federated algorithms, i.e. aggregation algorithms using balanced sampling, loss re-weighting and gradient tuning, could be deployed \cite{yang2021federated, shen2021agnostic}. Thirdly, future work could focus on individualizing the global model: The data owner could benefit from a model fine-tuned to local data.

\section{Conclusions}\label{sec:conclusion}

Federated learning is a cost and performance-effective approach for transcriptomic data analysis, enabling collaborative model training across distributed data sources without sharing raw sensitive data, while also alleviating computational burdens on individual clients. We have analyzed the challenges of applying federated learning to transcriptomic data regarding architectural heterogeneity, statistical heterogeneity, Gaussian noise and resource consumption. In particular, we tested two real-world data sets and use cases with varying numbers of training rounds and clients, and we compared a centralized baseline with two FL frameworks. \\

Finally, our findings confirm that FL is applicable and beneficial in precision medicine enabling cross-institutional and robust disease prognosis and cell type classification using transcriptomic data. While both FL frameworks have different strengths, both can be used to readily build federated models.

%

\bibliographystyle{elsarticle-num}
\bibliography{main}

\end{document}